\documentclass[10pt,conference]{IEEEtran}
\IEEEoverridecommandlockouts

\usepackage{amsmath,amssymb,amsfonts}
\usepackage{graphicx}
\usepackage{textcomp}
\usepackage{booktabs}
\usepackage{array}
\usepackage{hyperref}
\usepackage{tikz}
\usetikzlibrary{arrows.meta,positioning,shapes.geometric}

\begin{document}

\title{Directional Contextual Representations for Dependency Relations: Why Cross-Direction
Pairing Fails}

\author{
\IEEEauthorblockN{
\textit{Sai Krishna Arthanari},
\textit{JaeHyeong Chang},
\textit{Chengzhe Sun},
\textit{Siwei Lyu}
}
\IEEEauthorblockA{
Institute for Artificial Intelligence and Data Science (IAD), University at Buffalo\\
Buffalo, NY, USA\\
arthanarisaikrishna@gmail.com, jchang46@buffalo.edu, csun22@buffalo.edu, siweilyu@buffalo.edu
}
}

\maketitle

\begin{abstract}
Splitting a bidirectional LSTM's contextual representation into a forward-only $F_i$ (strictly a
function of tokens $1..i$) and a backward-only $B_i$ (strictly a function of tokens $i..n$) beats
either alone and beats a fused self-attention representation for dependency relation-type
classification. But a specific, natural extension of this idea -- pairing a token's forward state
against a \emph{candidate}'s backward state (``cross-direction'' pairing, $F_i$ vs.\ $B_j$) --
consistently \emph{underperforms} same-direction pairing, and the penalty \emph{grows}, not
shrinks, with token distance, both paired-bootstrap significant. We diagnose why using a
frozen-trunk methodology: architectural information leakage between directions is impossible by
construction (a single-layer BiLSTM, verified by code inspection); 93\% of the same-vs-cross gap
survives freezing the trunk and training only fresh heads, ruling out training-co-adaptation as
the primary cause; linear regression shows partial representational redundancy between $F_i$ and
$B_i$ ($R^2{=}0.324$ vs.\ $0.028$ for a shuffled control) and a linear probe shows partial
anticipatory encoding of upcoming tokens in $F_i$ (36.5\% vs.\ 17.2\% majority baseline) -- real
effects, but neither alone, nor combined, cleanly explains the full gap. Extended frozen-trunk
diagnostics (a positional probe and a distance-decay probe) show directional information is
genuinely stored but not exactly positioned, and propagates only a few tokens before decaying to
baseline -- consistent with, and mechanistically underneath, the distance-growth finding. We
validate the core claim three ways: a parameter-matched Transformer backbone (a fresh end-to-end
BiLSTM still significantly outperforms it, Cohen's $d$ shrinking from 0.216 to 0.085 but staying
significant); a second UD English treebank of a substantially different genre (GUM), on which both
the directional-splitting and cross-direction-failure findings replicate closely; and a literature
search that found no prior work performing this specific cross-direction ablation, reported
plainly as a novelty claim rather than assumed.
\end{abstract}


\section{Introduction}

Contextual encoders that process a sequence in both directions -- bidirectional LSTMs, and, less
explicitly, self-attention -- are ubiquitous in NLP, but the specific contribution of
\emph{directional} structure to relational tasks like dependency parsing is rarely isolated. Most
architectures fuse forward and backward (or all-to-all) information into a single representation
before any relational comparison happens, so it is not obvious whether keeping directions separate
through the comparison step itself matters, and if so, in which specific combinations. This paper
asks that question narrowly and empirically: does directional structure in a contextual
representation carry recoverable relational information, and specifically, is comparing a token's
state from \emph{one} direction against a candidate's state from the \emph{opposite} direction
informative, or not? The $O(n^2)$ cost of unrestricted self-attention \cite{vaswani2017attention}
is useful background for why representation structure is worth studying carefully, but reducing
that cost is not this paper's question -- we make no adaptive-computation or efficiency claim here.

We use a single-layer bidirectional LSTM trunk, whose two directions architecturally guarantee a
forward-only representation $F_i$ (strictly a function of tokens $1..i$) and a backward-only
representation $B_i$ (strictly a function of tokens $i..n$), with no attention or mixing layer
between the recurrence and the split -- a property we verify, not assume. On Universal
Dependencies English EWT, we test a sequence of specific, falsifiable claims and preview them here
so the results read as confirmation or refutation, not an undifferentiated list of numbers:
\textbf{(1)} does splitting into $F$/$B$ beat either alone or a fused self-attention baseline?
\textbf{(2)} does \emph{cross-direction} pairing -- comparing $F_i$ to a candidate's $B_j$ -- add
signal beyond same-direction pairing? \textbf{(3)} if it does not, why -- architectural leakage,
representational redundancy, a training artifact, or something else? \textbf{(4)} does the
resulting picture hold under real scrutiny: a parameter-matched alternative backbone, a second
dataset, and a check of the relevant literature?

\begin{itemize}
\item \textbf{(1) Yes, unfused $F{+}B$ wins.} Combining $F$ and $B$ beats either alone and beats a
fused self-attention representation for relation-type classification, though bootstrap validation
shows this specific margin is statistically robust but a small standardized effect.
\item \textbf{(2) No, cross-direction pairing loses, and loses more with distance.} $F_i$-vs-$B_j$
pairing is consistently the weakest pairwise construction tested, and its penalty relative to
same-direction pairing grows monotonically with token distance -- the opposite of a ``masked by
short-range dominance'' account -- both bootstrap-significant.
\item \textbf{(3) Partially diagnosed, not fully explained.} Architectural leakage is ruled out by
construction; the gap is mostly not a training-co-adaptation artifact (93\% survives freezing the
trunk); representational redundancy and anticipatory encoding are both real but partial effects,
and extended positional/distance probes show information is stored but imprecisely positioned and
short-range -- together consistent with, but not a complete mechanistic account of, the failure.
\item \textbf{(4) Holds under scrutiny.} A parameter-matched Transformer backbone narrows but does
not close a fresh BiLSTM's advantage; a second UD treebank of a different genre (GUM) replicates
both the directional-splitting and cross-direction-failure findings closely; and a literature
search found no prior work on this specific cross-direction ablation.
\end{itemize}

\section{Related Work}

\textbf{Self-attention.} Vaswani et al.\ \cite{vaswani2017attention} introduced scaled dot-product
self-attention, where every token attends to every other via one fused, all-to-all computation --
useful background for why keeping directions separate is a non-trivial design choice, though this
paper does not address attention's computational cost. \textbf{Bidirectional representations.}
Peters et al.\ \cite{peters2018deep} (ELMo) build contextualized word representations from a
bidirectional LSTM's internal states, combining forward and backward directions for transfer; we
instead keep $F_i$/$B_i$ separate through the pairwise-comparison step itself, to isolate which
directional combinations carry relational signal -- directly comparable prior art for the
splitting question, though ELMo does not test cross-direction pairwise comparison specifically.
\textbf{Biaffine parsing.} Dozat and Manning \cite{dozat2017deep} score candidate dependency arcs
with a biaffine function of two tokens' representations, trained with a per-sentence softmax over
candidate heads; our backbone-comparison arc scorer (Section~\ref{sec:results}) follows this same
structured-prediction recipe, on an explicit, hand-specified feature rather than a learned
biaffine transform, keeping the pairwise representation interpretable. \textbf{Universal
Dependencies.} Nivre et al.\ \cite{nivre2020universal} describe UD; we use English EWT as the
primary testbed and English GUM, a substantially different genre, for cross-dataset validation.

\textbf{Cross-direction pairing: a literature check.} We searched specifically for prior work
comparing a token's forward-direction state against another token's backward-direction state (or
symmetric variants) in a bidirectional recurrent representation, and for prior probing work
establishing an asymmetry between what forward and backward directions predict about neighboring
tokens. We found no directly relevant prior work under this or related framings. We report the
cross-direction pairwise ablation in this paper as, to our knowledge, novel, while noting plainly
that this search was not exhaustive.

\section{Methodology}

We deliberately use a single-layer bidirectional LSTM, not a Transformer, as the trunk throughout
the core ablation (Sections~\ref{sec:results}A--C), stated explicitly rather than left implicit.
The requirement is that $F_i$/$B_i$ be provably, architecturally disjoint -- a single-layer BiLSTM
gives this for free, verifiable by inspection. A Transformer does not: self-attention mixes every
position and direction from layer one, so extracted ``forward''/``backward'' views would need
independent verification against the very contamination this study isolates. We use the BiLSTM as
a clean-room diagnostic instrument, not a smaller Transformer stand-in; a fresh Transformer backbone
is used only as an external validation point (Section~\ref{sec:results}D), not as an alternative
substrate for the $F$/$B$-splitting ablation itself (Section~\ref{sec:limitations}).

\subsection{Forward, backward, and positional representations}

\begin{figure}[!t]
\centering
\resizebox{0.95\linewidth}{!}{%
\begin{tikzpicture}[
  node distance=6mm and 5mm,
  box/.style={draw, rounded corners, minimum height=7mm, minimum width=16mm, align=center, font=\footnotesize},
  tok/.style={draw, minimum height=5mm, minimum width=6mm, font=\scriptsize},
  myarrow/.style={-{Latex[length=2mm]}}
]
  \node[tok] (x1) {$x_1$};
  \node[tok, right=3mm of x1] (x2) {$x_2$};
  \node[font=\scriptsize, right=2mm of x2] (dots1) {$\cdots$};
  \node[tok, right=2mm of dots1] (xi) {$x_i$};
  \node[font=\scriptsize, right=2mm of xi] (dots2) {$\cdots$};
  \node[tok, right=2mm of dots2] (xn) {$x_n$};

  \node[box, below=of x2, xshift=11mm] (embed) {embedder\\(word+UPOS)};
  \node[box, below=of embed] (bilstm) {1-layer BiLSTM};

  \node[box, below left=of bilstm] (F) {$F_i$\\forward output};
  \node[box, below right=of bilstm] (B) {$B_i$\\backward output};
  \node[box, right=10mm of B] (P) {$P_i$\\sinusoidal (fixed)};

  \node[box, below=14mm of bilstm] (H) {$H_i=\mathrm{concat}(F_i,B_i,P_i)$};

  \draw[myarrow] (x1) -- (embed);
  \draw[myarrow] (x2) -- (embed);
  \draw[myarrow] (xi) -- (embed);
  \draw[myarrow] (xn) -- (embed);
  \draw[myarrow] (embed) -- (bilstm);
  \draw[myarrow] (bilstm) -- (F);
  \draw[myarrow] (bilstm) -- (B);
  \draw[myarrow] (F) -- (H);
  \draw[myarrow] (B) -- (H);
  \draw[myarrow] (P) -- (H);
\end{tikzpicture}%
}
\caption{Trunk architecture. A single-layer BiLSTM over token embeddings exposes $F_i$ and $B_i$
directly, with no attention or mixing layer between the LSTM and the split. $P_i$ is computed
independently of the LSTM. $H_i=\mathrm{concat}(F_i,B_i,P_i)$ is the per-token diagnostic
representation used by the positional and distance-decay probes (Section~\ref{sec:results}C).}
\label{fig:trunk}
\end{figure}
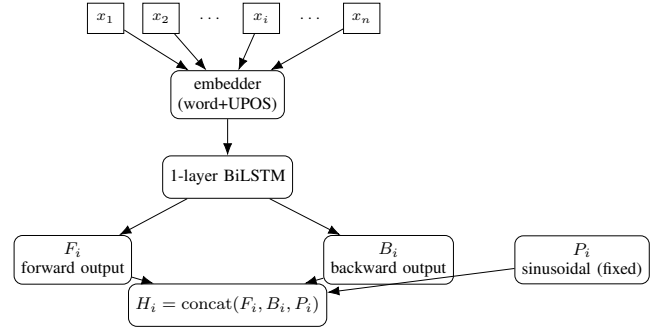

For a sentence of $n$ tokens we build a shared trunk: word embeddings (dimension 100, vocabulary
from training tokens with minimum frequency 2, plus \texttt{<unk>}/\texttt{<pad>}) concatenated
with UPOS embeddings (dimension 32), a 132-dimensional per-token input, fed through a single-layer
bidirectional LSTM (hidden size 128 per direction, \texttt{nn.LSTM(..., bidirectional=True)}, no
\texttt{num\_layers} override). The forward-direction hidden state $F_i \in \mathbb{R}^{128}$ and
backward-direction hidden state $B_i \in \mathbb{R}^{128}$ are sliced directly from the raw
bidirectional output with no attention or mixing layer in between:
\[
F_i = \mathrm{LSTM}_{\rightarrow}(x_1, \ldots, x_i), \qquad
B_i = \mathrm{LSTM}_{\leftarrow}(x_i, \ldots, x_n).
\]
Because the LSTM is single-layer and $F_i$/$B_i$ are read directly off its two output halves,
$F_i$ is by construction a strict function of tokens $1..i$ only, and $B_i$ of tokens $i..n$ only
-- no mechanism exists for information from the ``wrong'' side to leak into either representation,
which we rely on directly in the diagnostic analysis (Section~\ref{sec:results}C). $P_i \in
\mathbb{R}^{64}$ is the classic sinusoidal positional encoding, a fixed function of absolute index
$i$: $P_i[2k]=\sin(i/10000^{2k/64})$, $P_i[2k{+}1]=\cos(i/10000^{2k/64})$. Figure~\ref{fig:trunk}
summarizes the trunk.

\subsection{Pairwise representations tested}

For a candidate pair $(i,j)$ (an arbitrary ordered token pair, or a gold head-dependent pair for
Section~\ref{sec:results}A) we tested the following pairwise feature constructions $R_{ij}$, each
fed to an identically-shaped classifier head (single hidden layer, ReLU, dropout 0.2, linear
output) so comparisons isolate the representation, not classifier capacity:
{\footnotesize
\begin{align*}
R^{\mathrm{f\_only}}_{ij} &= [F_i, F_j, F_i{\odot}F_j, |F_i{-}F_j|] \\
R^{\mathrm{b\_only}}_{ij} &= [B_i, B_j, B_i{\odot}B_j, |B_i{-}B_j|] \\
R^{\mathrm{f\_plus\_b}}_{ij} &= [F_i, B_i, F_j, B_j, F_i{\odot}B_j, B_i{\odot}F_j, \\
&\quad |F_i{-}F_j|, |B_i{-}B_j|] \\
R^{\mathrm{full\_ref}}_{ij} &= R^{\mathrm{f\_plus\_b}}_{ij} \oplus (P_i - P_j) \\
R^{\mathrm{same\_f}}_{ij} &= R^{\mathrm{f\_only}}_{ij} \quad \text{(binary-task naming)} \\
R^{\mathrm{same\_b}}_{ij} &= R^{\mathrm{b\_only}}_{ij} \quad \text{(binary-task naming)} \\
R^{\mathrm{cross\_fb}}_{ij} &= [F_i, B_j, F_i{\odot}B_j, |F_i{-}B_j|] \\
R^{\mathrm{cross\_bf}}_{ij} &= [B_i, F_j, B_i{\odot}F_j, |B_i{-}F_j|] \\
R^{\mathrm{combined\_e}}_{ij} &= [F_i, B_j, B_i, F_j, P_i{-}P_j] \\
R^{\mathrm{ff\_plus\_fb\_concat}}_{ij} &= R^{\mathrm{same\_f}}_{ij} \oplus R^{\mathrm{cross\_fb}}_{ij} \\
R^{\mathrm{residual\_fb\_ff}}_{ij} &= R^{\mathrm{cross\_fb}}_{ij} - R^{\mathrm{same\_f}}_{ij}
\end{align*}}
where $\odot$ is elementwise product, $\oplus$ concatenation. $R^{full\_ref}$
($f\_plus\_b\_plus\_p$) is the strongest variant in Section~\ref{sec:results}A and is reused
unchanged as the feature formula for the full-pairwise arc scorer in the backbone comparison
(Section~\ref{sec:results}D). In $R^{residual\_fb\_ff}$ the first 128 dims ($F_i-F_i$) cancel to
zero by construction; the remaining 384 carry the signal. As a fused-representation baseline we
also trained a small (2-layer, 4-head, $d_{model}{=}256$) Transformer encoder over the same
embeddings (sinusoidal position added elementwise, standard practice), producing one fused $H_i$
per token with no forward/backward split: $R^{standard\_attention}_{ij} = [H_i, H_j,
H_i{\odot}H_j, |H_i{-}H_j|]$.

\subsection{Per-token diagnostic representation}

To probe what a token's own state encodes about position and neighboring tokens -- independent of
any specific comparison pair -- we use
\[
H_i = \mathrm{concat}(F_i, B_i, P_i) \in \mathbb{R}^{320}.
\]
$H_i$ is used here solely as the input to two frozen-trunk diagnostic probes
(Section~\ref{sec:results}C): a positional probe (does $H_i$, or its $F_i$/$B_i$/$F_i{+}B_i$
sub-parts, encode absolute position, and how exactly) and a distance-decay probe (how far does
directional information about specific neighboring tokens propagate). These probes characterize
what the representations store; we make no claim here about using $H_i$ for selective or adaptive
computation.

\subsection{Backbone comparison: BiLSTM vs.\ Transformer, parameter-matched}

To test whether the $F$/$B$-splitting advantage (Section~\ref{sec:results}A, a classification
task) extends to a genuine structured-prediction task, we trained full-pairwise arc scorers for
unlabeled dependency head-finding: $\mathrm{score}(i,j) = \mathrm{MLP}(R^{full\_ref}_{ij})$,
trained end-to-end (trunk and scorer optimized jointly) with a per-sentence softmax
cross-entropy over candidate heads $j$ in the same sentence, following the same
structured-prediction recipe as \cite{dozat2017deep}. We compare three backbones under this
identical recipe: the BiLSTM trunk above; a Transformer backbone (the \texttt{standard\_attention}
trunk from Section~\ref{sec:results}A's fused baseline); and, since that Transformer has
$\sim$52\% more parameters than the BiLSTM -- a real confound for a backbone comparison -- a
\emph{parameter-matched} Transformer ($d_{model}{=}144$, 2 layers, 4 heads, $\dim_{ff}{=}288$)
reduced until its total parameter count sits within 3\% of the BiLSTM's.

\section{Experimental Setup}

All experiments use Universal Dependencies English EWT (\texttt{en\_ewt-ud-\{train,dev,test\}
.conllu}) as the primary testbed, with English GUM (a substantially different genre: academic,
fiction, how-to, news, interview, and travel-guide text, vs.\ EWT's informal web/blog/email/review
text) for cross-dataset validation (Section~\ref{sec:results}E). Both are parsed with a minimal
hand-written CoNLL-U parser skipping comment lines and multiword/empty-node IDs (\texttt{-}
/\texttt{.} in the ID), retaining per token its form, UPOS, integer head index, and DEPREL,
reduced to its base label (before any \texttt{:}); the 15 most frequent training-split base labels
become classes, the rest map to \texttt{other} (16 classes). Arcs with \texttt{HEAD==0} (virtual
root) are excluded throughout. EWT splits: 12,544/2,001/2,077 sentences, 192,034/23,147/23,017
arcs. GUM splits: 11,314/1,575/1,464 sentences, 188,909/26,933 train/test arcs.

Training protocol, held fixed across the study for comparability: AdamW, learning rate $10^{-3}$,
weight decay $10^{-5}$; batch size 32 sentences; 8 epochs; classifier head shape
$\mathrm{Linear}(\mathrm{pair\_dim} \rightarrow 256) \rightarrow \mathrm{ReLU} \rightarrow
\mathrm{Dropout}(0.2) \rightarrow \mathrm{Linear}(256 \rightarrow \mathrm{num\_classes})$; seeds
42--44 as the default (3 seeds per configuration), extended to seeds 42--46 (5 seeds) for the two
most load-bearing claims -- the relation-classification ablation (Section~\ref{sec:results}A) and
the backbone comparison (Section~\ref{sec:results}D). Hardware: a single NVIDIA RTX 4090 (24GB).
The binary edge-existence task (Sections~\ref{sec:results}B--C) uses a fixed 1:1 positive/negative
dataset (gold arcs vs.\ randomly sampled non-arc index pairs per sentence, negative-sampling seed
42, reused unchanged across variants and training seeds so comparisons isolate model variance, not
data variance).

\section{Results}
\label{sec:results}

Four questions, traced in order: (A) does $F$/$B$ splitting beat fused? (B--C) does cross-direction
pairing add signal, and why does it fail? (D--E) does this hold under real scrutiny?

\subsection*{A. Does splitting $F$/$B$ beat either alone or fused?}

Table~\ref{tab:phase1} reports dependency relation-type classification (16-way, gold arcs) test
accuracy and macro-F1 for each pairwise representation (single seed 42), plus a 5-seed (42--46)
replication of four variants confirming the single-seed figures.

\begin{table}[!t]
\centering
\caption{Relation-type classification, test set. Single-seed columns are the primary run; 5-seed
column (mean$\pm$std, seeds 42--46) confirms it.}
\label{tab:phase1}
\begin{tabular}{@{}lcccc@{}}
\toprule
Variant & Test Acc.\ & 5-seed Acc.\ & Macro-F1 & Params \\
\midrule
f\_only & 0.9667 & 0.9650$\pm$.0016 & 0.9654 & 1.39M \\
b\_only & 0.9625 & 0.9622$\pm$.0011 & 0.9617 & 1.39M \\
f\_plus\_b & 0.9725 & -- & 0.9720 & 1.52M \\
\textbf{f\_plus\_b\_plus\_p} & \textbf{0.9733} & \textbf{0.9716$\pm$.0020} & \textbf{0.9732} & 1.54M \\
standard\_attention & 0.9652 & 0.9646$\pm$.0009 & 0.9645 & 2.34M \\
\bottomrule
\end{tabular}
\end{table}

Combining unfused $F$ and $B$ gives the largest jump over either direction alone, beating fused
self-attention by $\sim$0.7 points with $\sim$35\% fewer parameters. Paired bootstrap (seed 42,
5000 resamples, $n{=}23{,}017$) confirms \texttt{f\_plus\_b\_plus\_p} beats \texttt{f\_only}
(+0.0080, 95\% CI $(+0.0059,+0.0101)$, Cohen's $d{=}0.049$), \texttt{b\_only} (+0.0106, CI
$(+0.0083,+0.0127)$, $d{=}0.064$), and \texttt{standard\_attention} (+0.0076, CI
$(+0.0054,+0.0097)$, $d{=}0.045$) -- all significant, all small standardized effects. This
establishes the starting point: directional splitting works. The rest of the paper asks the
sharper question of \emph{which} directional combinations carry the signal.

\subsection*{B. Does cross-direction pairing add signal?}

Table~\ref{tab:phase2} reports binary edge-existence detection (is there a dependency arc between
$i$ and $j$ at all), mean $\pm$ std over 3 seeds, on a 1:1 balanced gold-arc-vs.-random-pair
dataset.

\begin{table}[!t]
\centering
\caption{Binary edge existence, test set, mean $\pm$ std over 3 seeds.}
\label{tab:phase2}
\begin{tabular}{@{}lccc@{}}
\toprule
Variant & Test Acc.\ & Test F1 & Params \\
\midrule
same\_f (FF) & 0.9362$\pm$0.0004 & 0.9364$\pm$0.0003 & 1.39M \\
same\_b (BB) & 0.9351$\pm$0.0014 & 0.9354$\pm$0.0015 & 1.39M \\
cross\_fb (FB) & 0.9050$\pm$0.0018 & 0.9058$\pm$0.0025 & 1.39M \\
cross\_bf (BF) & 0.8588$\pm$0.0022 & 0.8582$\pm$0.0025 & 1.39M \\
\textbf{combined\_e} & \textbf{0.9543$\pm$0.0004} & \textbf{0.9547$\pm$0.0004} & 1.40M \\
full\_ref & 0.9522$\pm$0.0015 & 0.9524$\pm$0.0016 & 1.54M \\
\bottomrule
\end{tabular}
\end{table}

Cross-direction pairing is not an additional source of signal -- it is the clear \emph{loser}:
\texttt{cross\_fb} trails \texttt{same\_f}/\texttt{same\_b} by $\sim$3 points and
\texttt{cross\_bf} trails by $\sim$7--8 points, gaps 30--80$\times$ larger than seed-to-seed noise
($\mathrm{std}\le0.0025$ throughout). Paired bootstrap (seed 42, sentence-independent example
resampling, 5000 iterations) confirms \texttt{cross\_fb} trails \texttt{same\_f} by $-$0.0308
(95\% CI $(-0.0336,-0.0279)$, Cohen's $d{=}-0.100$) -- significant, small standardized effect. The
best variant overall is \texttt{combined\_e}, a raw concatenation with no products or absolute
differences, not a crafted interaction term.

Does this weakness concentrate at short range, where a same-direction shortcut might dominate the
aggregate, masking a real long-range cross-direction advantage? Table~\ref{tab:phase3} tests this
directly by token distance $|i-j|$.

\begin{table}[!t]
\centering
\caption{cross\_fb $-$ same\_f accuracy/F1 gap by token distance (test, mean over 3 seeds).}
\label{tab:phase3}
\begin{tabular}{@{}lccccc@{}}
\toprule
$|i-j|$ bucket & 1--2 & 3--5 & 6--10 & 11--20 & 21+ \\
\midrule
$n$ (test pairs) & 19,571 & 11,639 & 7,576 & 5,188 & 1,922 \\
Accuracy gap & $-$0.0233 & $-$0.0300 & $-$0.0384 & $-$0.0502 & $-$0.0538 \\
F1 gap & $-$0.0166 & $-$0.0324 & $-$0.0606 & $-$0.1045 & $-$0.1137 \\
\bottomrule
\end{tabular}
\end{table}

\begin{figure}[!t]
\centering
\includegraphics[width=0.98\linewidth]{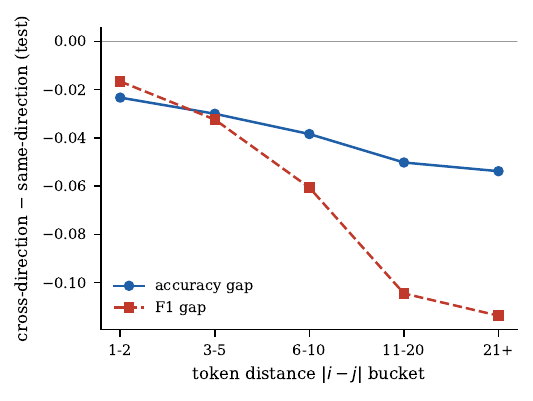}
\caption{cross\_fb $-$ same\_f accuracy and F1 gap by token-distance bucket (test set, mean over 3
seeds). Both gaps are monotonically negative and grow with distance -- the opposite of the
``masked by short-range dominance'' prediction.}
\label{fig:phase3}
\end{figure}

The cross-direction penalty grows monotonically with distance -- nearly 7$\times$ larger in F1 at
21+ than at 1--2 -- the opposite of the ``masked by short-range dominance'' account.
Figure~\ref{fig:phase3} plots both gaps. Paired bootstrap on the difference of gaps (gap at 21+
minus gap at 1--2, stratified resampling within each bucket, 5000 iterations) gives $-$0.0371
(95\% CI $(-0.0516,-0.0227)$) -- the growth trend itself is statistically robust, not just
numerically monotonic.

\subsection*{C. Why does cross-direction pairing fail?}

We probe four candidate explanations on one frozen trunk (\texttt{combined\_e}, seed 42, zero
further training). \textbf{(a) Architectural leakage} is ruled out by code inspection alone
(Section~\ref{sec:results}: the trunk is single-layer with no mixing before the $F$/$B$ split; see
also Methodology). \textbf{(b) Representational redundancy:} linear regression $F_i \to B_i$
achieves $R^2{=}0.324$ out-of-sample vs.\ $R^2{=}0.028$ for a shuffled-partner control
(11.5$\times$ higher, but far from the $R^2{\to}1$ full collapse would require). \textbf{(c)
Training-dynamics artifact:} fresh classifier heads trained on the frozen trunk reproduce
\textbf{93\%} of the end-to-end same\_f-vs-cross\_fb gap (2.96 of 3.18 points; Table~\ref{tab:phase4}),
largely ruling out co-adaptation as the primary cause. \textbf{(d) Anticipatory encoding:} $F_i$
predicts the \emph{next} token's UPOS at 36.5\% accuracy (majority baseline 17.2\%, same-token
ceiling 98.1\%) -- real but partial anticipatory signal, and not, on its own, an explanation for
why cross-direction pairing is actively \emph{worse} rather than merely no-better.

\begin{table}[!t]
\centering
\caption{Frozen-trunk vs.\ end-to-end same\_f$-$cross\_fb accuracy gap.}
\label{tab:phase4}
\begin{tabular}{@{}lcc@{}}
\toprule
Condition & same\_f Acc.\ & cross\_fb Acc.\ \\
\midrule
Frozen trunk (fresh heads, 3 seeds) & 0.9240$\pm$0.0008 & 0.8943$\pm$0.0006 \\
End-to-end & 0.9367$\pm$0.0001 & 0.9049$\pm$0.0017 \\
\bottomrule
\end{tabular}
\end{table}

\textbf{Extended diagnostic 1: is position stored, and is it exact?} Does the LSTM recurrence
already implicitly encode absolute position, making $P_i$ redundant, or does $P_i$ add real
recoverable precision? We probe position (decile bucket, 10 classes; and normalized $i/(n{-}1)$,
regression) from four sources on the frozen trunk (Table~\ref{tab:probepos}), each with both an
MLP head (as used throughout this section) and a pure linear head (no hidden layer), to test
whether the encoded information is linearly accessible.

\begin{table}[!t]
\centering
\caption{Position-probe bucket accuracy, MLP vs.\ linear head, mean over 3 seeds (majority
baseline 0.1367). MAE columns are for the MLP head's normalized-position regression (median
baseline MAE 0.2718).}
\label{tab:probepos}
\begin{tabular}{@{}lcccc@{}}
\toprule
Source & Dim & MLP Acc.\ & Linear Acc.\ & MLP Frac.\ MAE \\
\midrule
$F_i$ & 128 & 0.3310 & 0.3103 & 0.1920 \\
$B_i$ & 128 & 0.3314 & 0.2983 & 0.1919 \\
$F_i{+}B_i$ (no $P$) & 256 & 0.4830 & 0.4072 & 0.1436 \\
\textbf{$H_i$ (with $P_i$)} & 320 & \textbf{0.6679} & \textbf{0.5140} & \textbf{0.1350} \\
\bottomrule
\end{tabular}
\end{table}

Neither answer is clean. The recurrence already encodes real position without any explicit signal
($F{+}B$ reaches $\sim$3$\times$ the majority baseline with no $P_i$ at all under the MLP head),
so $P_i$ is not filling a total void -- but $P_i$ is not redundant either: $H_i$ jumps a further
$\sim$19 points over $F{+}B$ alone (MLP head). The linear head always trails the MLP, sometimes
narrowly (direction-adjacent sources) and sometimes by a lot ($H_i$: $-$15.4 points, the largest
linear-vs-MLP gap we observed) -- the position information exists but is not fully linearly
accessible, especially once $P_i$'s sinusoidal encoding is involved, since decoding it into a
discrete bucket is an inherently nonlinear operation.

\textbf{Extended diagnostic 2: how far does directional information propagate?} We predict
$\mathrm{UPOS}(i{+}\Delta)$ from $F_i$ alone (symmetrically $\mathrm{UPOS}(i{-}\Delta)$ from
$B_i$) for $\Delta\in\{1,2,3,5,10,20\}$, against a majority baseline and a same-token
direct-observation ceiling ($F_{i+\Delta}\to\mathrm{UPOS}(i{+}\Delta)$, $\approx$0.98 throughout,
confirming the probe architecture is not the bottleneck; Table~\ref{tab:probedist}).

\begin{table}[!t]
\centering
\caption{Distance-decay probe accuracy vs.\ majority baseline, frozen trunk, seed 42.}
\label{tab:probedist}
\begin{tabular}{@{}rcccc@{}}
\toprule
$\Delta$ & Fwd.\ probe & Fwd.\ maj.\ & Bwd.\ probe & Bwd.\ maj.\ \\
\midrule
1 & 0.3669 & 0.1721 & 0.3518 & 0.1722 \\
2 & 0.2240 & 0.1729 & 0.1985 & 0.1543 \\
3 & 0.1889 & 0.1764 & 0.1686 & 0.1552 \\
5 & 0.1814 & 0.1814 & 0.1541 & 0.1540 \\
10 & 0.1806 & 0.1847 & 0.1549 & 0.1532 \\
20 & 0.1797 & 0.1805 & 0.1457 & 0.1535 \\
\bottomrule
\end{tabular}
\end{table}

Both directions decay sharply, hitting their majority baseline by $\Delta{=}5$ (forward) or
$\Delta{=}3$ (backward) -- directional information about a specific token's category propagates
only a handful of positions before vanishing into noise, a concrete mechanistic number underneath
part B's task-level distance-growth finding: cross-direction pairing has less and less real signal
to draw on as $|i-j|$ grows, exactly where its penalty grows largest.

\subsection*{D. Does this hold under a fresh, parameter-matched backbone?}

Table~\ref{tab:backbone} compares three full-pairwise arc scorers trained end-to-end (trunk and
scorer jointly) for real unlabeled dependency head-finding (UAS), 5 seeds each.

\begin{table}[!t]
\centering
\caption{Full-pairwise arc scorer backbone comparison, mean $\pm$ std over 5 seeds.}
\label{tab:backbone}
\begin{tabular}{@{}lccc@{}}
\toprule
Backbone & Params & Dev UAS & Test UAS \\
\midrule
\textbf{BiLSTM} & 1.535M & \textbf{0.8580$\pm$.0012} & \textbf{0.8609$\pm$.0014} \\
Transformer (unmatched) & 2.339M & 0.8308$\pm$.0047 & 0.8305$\pm$.0043 \\
Transformer (matched) & 1.490M & 0.8337$\pm$.0029 & 0.8349$\pm$.0037 \\
\bottomrule
\end{tabular}
\end{table}

Parameter-matching narrows the gap (test UAS $+3.04$ points unmatched $\to$ $+2.60$ points
matched) but does not close it. Table~\ref{tab:bootback} reports paired bootstrap (seed 42,
sentence-level resampling -- whole sentences, not tokens -- 5000 iterations) for both
comparisons.

\begin{table}[!t]
\centering
\caption{Paired bootstrap, BiLSTM $-$ Transformer UAS (seed 42, sentence-level).}
\label{tab:bootback}
\begin{tabular}{@{}lccc@{}}
\toprule
Comparison & Point & 95\% CI & Cohen's $d$ \\
\midrule
BiLSTM $-$ Transf.\ (unmatched) & +0.0386 & (+0.0331, +0.0444) & 0.216 \\
BiLSTM $-$ Transf.\ (matched) & +0.0234 & (+0.0180, +0.0287) & 0.085 \\
\bottomrule
\end{tabular}
\end{table}

Both CIs exclude zero: at near-identical parameter count, the BiLSTM still significantly
outperforms the Transformer on this task, though the standardized effect shrinks from moderate
($d{=}0.216$) to small ($d{=}0.085$). This extends part A's classification-only finding to a real
structured-prediction task under a fair parameter budget.

\subsection*{E. Does this hold on a different genre?}

Table~\ref{tab:gum} replicates the two core claims (parts A and B) on UD English GUM (academic,
fiction, how-to, news, interview, and travel-guide text -- substantially different from EWT's
informal web/blog/email/review text; 11,314/1,575/1,464 train/dev/test sentences, 188,909/26,933
train/test arcs, comparable scale to EWT, identical protocol and seeds).

\begin{table}[!t]
\centering
\caption{EWT vs.\ GUM, core claims (test set).}
\label{tab:gum}
\begin{tabular}{@{}lcc@{}}
\toprule
Claim & EWT & GUM \\
\midrule
$f\_plus\_b - $standard\_attn.\ (acc.\ gap) & +0.0073 & +0.0077 \\
same\_f $-$ cross\_fb (3-seed acc.\ gap) & +0.0312 & +0.0397 \\
\bottomrule
\end{tabular}
\end{table}

Both claims replicate closely on a substantially different genre: bidirectional splitting still
beats fused attention (GUM's full 5-variant ordering matches EWT's:
\texttt{f\_plus\_b\_plus\_p}$>$\texttt{f\_plus\_b}$>$\texttt{f\_only}$>$\texttt{standard\_attention}$>$
\texttt{b\_only}), and cross-direction pairing still loses by a comparable, if slightly larger,
margin. A third UD treebank in a different \emph{language} was not attempted, named here as an
open question rather than skipped silently (Section~\ref{sec:limitations}).

\section{Discussion}

\textbf{What replicated.} Directional splitting is genuinely informative and generalizes across
genre (part A, part E): unfused $F{+}B$ beats either alone and beats fused self-attention, and the
same ordering holds on GUM. Cross-direction pairing's failure is equally robust: consistently the
weakest construction, its penalty growing (not concentrating) with distance, both
bootstrap-confirmed and replicating on GUM. The backbone comparison (part D) shows this is not an
artifact of the specific backbone either -- a fresh, end-to-end, parameter-matched Transformer
still loses to the BiLSTM on a real structured-prediction task.

\textbf{What remains only partially explained.} Part C rules out architectural leakage cleanly and
training-co-adaptation largely (93\% of the gap survives trunk-freezing), but the positive
mechanism is incomplete: representational redundancy ($R^2{=}0.324$) and anticipatory encoding
(36.5\% vs.\ 17.2\%) are both real, above-control effects, yet neither in isolation -- nor, by
construction, obviously in combination -- accounts for why cross-direction pairing is actively
\emph{worse} than same-direction pairing rather than merely no-better. The extended positional and
distance-decay probes sharpen the picture (information is stored but imprecisely positioned, and
propagates only 3--5 tokens) without fully closing this gap. We consider the mechanism partially,
not fully, diagnosed, and say so plainly rather than overclaim a tidy causal story.

\textbf{On novelty.} We found no prior work performing this specific cross-direction pairwise
ablation, and report it as such -- a claim we hold loosely, since our literature search was
targeted, not exhaustive.

\section{Limitations}
\label{sec:limitations}

Two datasets, both English (EWT and GUM); a third treebank in a different language was not
attempted, an explicit open question rather than a silent gap. The core $F$/$B$-splitting ablation
(parts A--C) uses a single architecture family, a single-layer bidirectional LSTM, chosen because
it makes $F_i$/$B_i$ separation architecturally exact; the Transformer backbone in part D is used
only as an external validation point on a downstream task, not as an alternative substrate for
testing directional splitting itself -- whether a Transformer-derived forward/backward split
(e.g.\ via causal masking or paired unidirectional models) would show the same cross-direction
failure is untested. The literature search underlying our novelty claim (cross-direction pairwise
ablation) was targeted at specific query framings, not exhaustive. The mechanism behind
cross-direction pairing's failure remains partially, not fully, diagnosed: redundancy and
anticipatory encoding are real contributing factors but do not, individually or combined, fully
account for the size of the gap.

\clearpage

\section{Conclusion}

Splitting a bidirectional LSTM's representation into forward-only and backward-only parts is
genuinely useful for dependency relation discovery, beating both single-direction and fused
alternatives, and this generalizes across genre. But a specific, natural extension --
cross-direction pairwise comparison -- consistently fails, and fails more as token distance grows,
both statistically robust findings. We diagnosed this failure using a frozen-trunk methodology:
architectural leakage is impossible by construction, the gap is mostly not a training artifact,
and partial representational redundancy and anticipatory encoding are real contributing factors
that do not, individually, fully explain it -- a mechanism we report as partially, not fully,
understood. The core claims hold under a parameter-matched alternative backbone and a
substantially different-genre dataset, and, to our knowledge, this specific cross-direction
ablation has not been reported before. We present this as a focused, honestly-scoped empirical
result rather than a general theory of directional representations.

\end{document}